\documentclass[letterpaper]{article} %
\usepackage{aaai25}  %
\usepackage{times}  %
\usepackage{helvet}  %
\usepackage{courier}  %
\usepackage[hyphens]{url}  %
\usepackage{graphicx} %
\usepackage{natbib}  %
\usepackage{caption} %
\usepackage{newfloat}
\usepackage{listings}
\usepackage{fancyhdr}
\usepackage[hidelinks]{hyperref}
\DeclareCaptionStyle{ruled}{labelfont=normalfont,labelsep=colon,strut=off} %

\usepackage{xspace}
\newcommand{\name}{\mbox{NL2Plan}\xspace}

\nocopyright

\title{
    \name: Robust LLM-Driven Planning from Minimal Text Descriptions
}
\author {
    Elliot Gestrin,
    Marco Kuhlmann,
    Jendrik Seipp
}
\affiliations {
    Linköping University, Sweden\\
    firstname.lastname@liu.se
}

\usepackage{microtype}
\usepackage[main=british]{babel}
\useshorthands*{"}
\defineshorthand{"-}{\babelhyphen{hard}}
\defineshorthand{"=}{\babelhyphen{–}}
\defineshorthand{"/}{\babelhyphen{/}}

\usepackage{tikz}

\usepackage{tabularx}
\usepackage{multirow,makecell}
\usepackage{rotating}
\usepackage{booktabs}
\usepackage{varwidth} %

\def\baseline{\mbox{PDDL"-0}} %
\newcolumntype{R}{>{\raggedleft\arraybackslash}X}

\newcommand\predicate[1]{\texttt{#1}}
\newcommand\action[1]{\texttt{#1}}
\newcommand\type[1]{\texttt{#1}}
\newcommand\keyword[1]{\texttt{#1}}

\usepackage{enumitem}
\usepackage{import}
\usepackage[most]{tcolorbox}
\newcommand{\inlinecite}[1]{\citet{#1}}
\newcommand{\egcite}[1]{\citep[e.g.,][]{#1}}

\newcommand{\blocksworld}{\mbox{Blocksworld}\xspace}
\newcommand{\household}{\mbox{Household}\xspace}
\newcommand{\dimicy}{\mbox{DiMiCy}\xspace}
\newcommand{\dungeon}{\mbox{Dungeoncrawl}\xspace}
\newcommand{\robility}{\mbox{Robility}\xspace}
\newcommand{\rodrings}{Rod"-\mbox{Rings}\xspace}  %
\newcommand{\splitfish}{\mbox{Splitfish}\xspace}

\begin{document}

\maketitle
\thispagestyle{fancy}

\begin{abstract}
    Classical planners are powerful systems, but modeling tasks in input formats such as PDDL is tedious and error-prone. In contrast, planning with Large Language Models (LLMs) allows for almost any input text, but offers no guarantees on plan quality or even soundness. In an attempt to merge the best of these two approaches, some work has begun to use LLMs to automate parts of the PDDL creation process. However, these methods still require various degrees of expert input or domain-specific adaptations. We present \name, the first fully automatic system for generating complete PDDL tasks from minimal natural language descriptions. \name uses an LLM to incrementally extract the necessary information from the short text input before creating a complete PDDL description of both the domain and the problem which is finally solved by a classical planner. We evaluate \name on seven planning domains, five of which are novel and thus not in the LLM training data, and find that \name outperforms directly generating the files with an LLM+validator combination. As such, \name is a powerful tool for assistive PDDL modeling and a step towards solving natural language planning task with interpretability and guarantees.
\end{abstract}

\begin{figure}[tbp]
    \centering
    \includegraphics[width=\columnwidth, height=0.75\textheight, keepaspectratio]{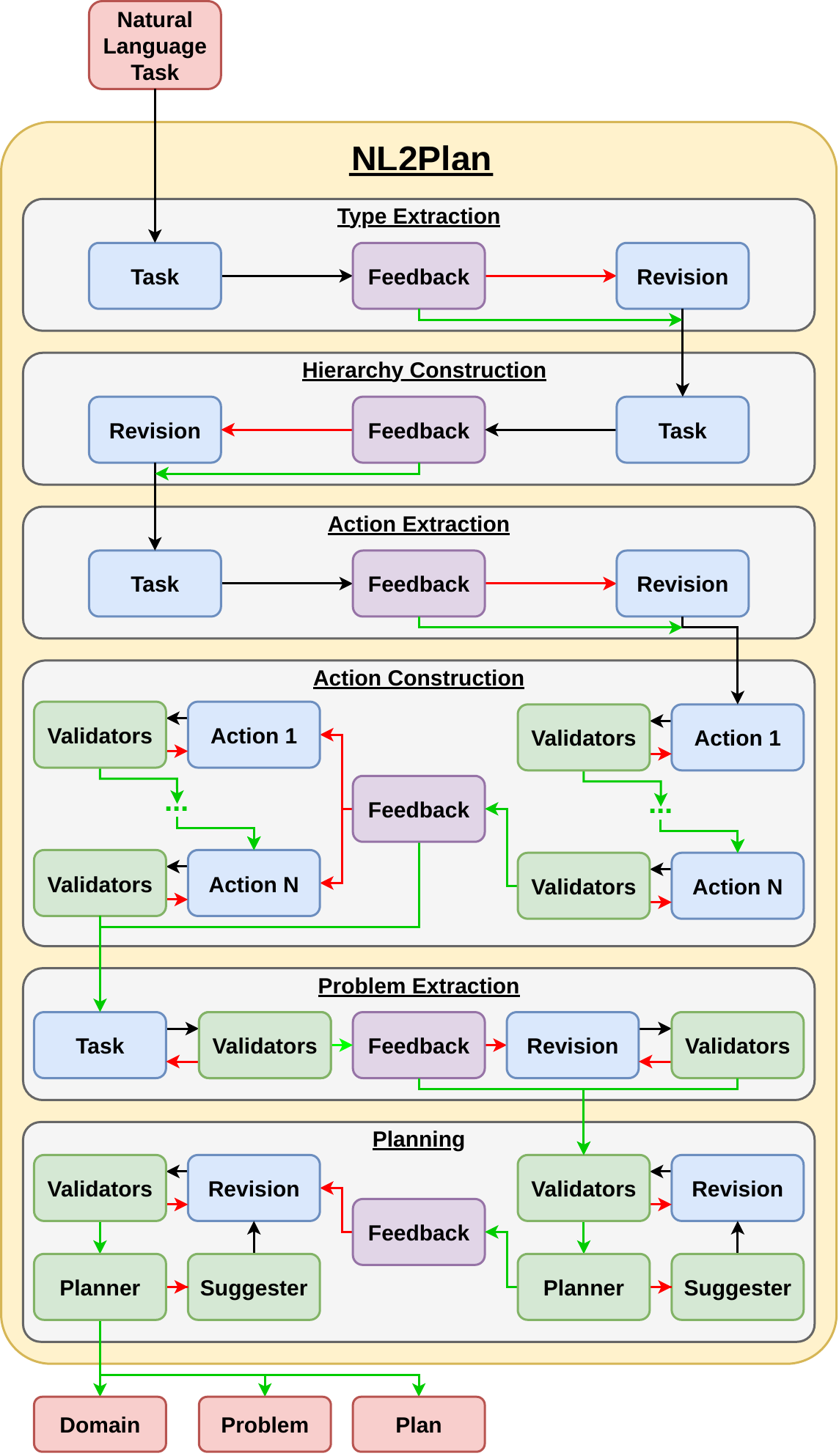}
    \caption{A block scheme of the six \name
    steps. During the Type Extraction step, we generate a set of object types,
    which are then structured into an inheritance tree by the Type Hierarchy
    step. Following this, we create a list of natural language action
    descriptions in the Action Extraction step, which are then formalized into a
    PDDL domain during the Action Construction step. During the Problem
    Extraction step, the initial and goal states are specified. Lastly, the
    Planning step involves a classical planner that attempts to solve the
    planning task with an LLM that revises the domain and problem if it proves
    to be unsolvable. Several steps include feedback, either from an LLM or a
    human, for semantic validation in addition to automatic validation tools for
    syntactical validation. Note that each validation loop can only occur at
    most a fixed number of times before progressing to the next substep with
    invalid code, as to avoid infinite loops. The user only has to interact with
    \name to provide the natural language problem and act on the plan.}
    \label{fig:method}
\end{figure}

\section{Introduction}

The field of AI planning \cite{ghallab-et-al-2004} has developed powerful domain-independent planning systems such as Fast Downward \cite{helmert-jair2006}, which are capable of receiving task descriptions and solving these efficiently.
The most common format for specifying such problems is the Problem Domain Definition Language (PDDL) \cite{haslum-et-al-2019}.
Modeling planning tasks is, however, a labor"-intensive task that requires PDDL"-trained users and an understanding of the domain under consideration.

In contrast, using large language models (LLMs) generally requires no training, only an understanding of the task and the ability to express it. For this reason, LLMs have been used to generate parts of PDDL models, such as the actions and predicates \cite{llmConstructWorldModels}, just the goal state \cite{lyu-etal-2023-faithful, xie2023translating}, or both the initial and goal state \cite{llmPlusP, dynamicPlanningLLM, LLMsOutOfDistribution}.

In this paper, we present \name, the first\footnote{We discuss caveats regarding \inlinecite{smirnov2024generatingconsistentpddldomains} in Section~\ref{sec:related}.} fully automated system capable of generating entire PDDL descriptions based on only a few sentences of natural language, without needing an expert user or any domain"-dependent adaptations.
While parts of \name build upon existing systems for obtaining partial PDDL models  from natural language \egcite{llmConstructWorldModels}, \name{} adds PDDL"-specific reasoning steps, automated common-sense feedback and planner"-driven correction suggestions.
As such, \name assists even PDDL"-trained users when modeling tasks
and it can often even directly find plans without user intervention.

We evaluate \name on seven planning domains, five of which are novel and thus not in the LLM's training data. We compare \name to a baseline method that uses an LLM and a syntax validator to directly write PDDL code. \name outperforms the baseline on all domains except for Blocksworld, where the baseline shows clear signs of memorization. \name models several tasks nearly perfectly, requiring a single change or fewer in 27.8\% of cases compared to the baseline which only gets equally close in 8.3\% of cases. As such, \name acts as a powerful assistive modeling tool.

Once the PDDL files are generated, either fully by \name or after minor user intervention, they can be passed to a classical planner which can find plans for the modeled task with several important guarantees, such as soundness and completeness. Used independently, \name models 260\% more tasks perfectly than the baseline. These tasks would give a correct and ideal plan when solved. This highlights how NL2Plan is a large step towards verifiable, interpretable and scalable planning for natural language tasks.

 All code for \name{}, its prompts, and evaluation data are available online \cite{gestrin-et-al-zenodo2025} and the latest version is maintained at {\color{blue}\href{https://github.com/mrlab-ai/NL2Plan}{https://github.com/mrlab-ai/NL2Plan}}.

\section{Background}
In this section, we introduce PDDL and describe the prompting techniques used for \name.

\newcommand{\truck}{\ensuremath{\mathit{truck}}}

\subsection{Planning Domain Definition Language (PDDL)}

\newcommand{\pddlDomain}{\ensuremath{D}}
\newcommand{\pddlProblem}{\ensuremath{P}}
\newcommand{\pddlTypes}{\ensuremath{T}}
\newcommand{\pddlType}{\ensuremath{t}}
\newcommand{\pddlPredicates}{\ensuremath{\mathcal{P}}}
\newcommand{\pddlPredicate}{\ensuremath{p}}
\newcommand{\pddlActions}{\ensuremath{A}}
\newcommand{\pddlAction}{\ensuremath{a}}
\newcommand{\pddlObjects}{\ensuremath{O}}
\newcommand{\pddlObject}{\ensuremath{o}}
\newcommand{\pddlInit}{\ensuremath{I}}
\newcommand{\pddlGoal}{\ensuremath{G}}

PDDL \cite{haslum-et-al-2019} is the predominant specification language for deterministic planning problems.
We assume basic familiarity with the semantics of PDDL and refer to the language definition for details \cite{fox-long-jair2003}.
We consider ``level 1'' of PDDL version 2.1, plus action costs.
This fragment includes all ADL features \cite{pednault-kr1989}, such as quantified and conditional effects, and negation, disjunction and quantification in conditions.

A PDDL task consists of a domain description \pddlDomain\ and a problem specification \pddlProblem. The domain specifies the common properties of the environment: the type hierarchy \pddlTypes, predicate set \pddlPredicates, and action schemas \pddlActions; $\pddlDomain = \langle \pddlTypes, \pddlPredicates, \pddlActions \rangle$. The problem
defines the task specifics: the available objects \pddlObjects, initial state \pddlInit, and the goal description \pddlGoal; $\pddlProblem = \langle \pddlObjects, \pddlInit, \pddlGoal \rangle$.

Within the domain $\pddlDomain = \langle \pddlTypes, \pddlPredicates, \pddlActions \rangle$, the type hierarchy \pddlTypes\ specifies which types exist and their relationship.
Each predicate $\pddlPredicate \in \pddlPredicates$
is defined over a tuple of arguments, where each argument has a type $t \in \pddlTypes$. By substituting all arguments of a predicate \pddlPredicate\ with objects $\pddlObject \in \pddlObjects$ of valid types, we obtain a Boolean proposition $\pddlPredicate(\pddlObject_1, \ldots , \pddlObject_m)$. We call this substitution process \emph{grounding} and the resulting proposition \emph{grounded}.

Similarly, \pddlActions\ is a set of action schemas $\pddlAction \in \pddlActions$, each of which in turn is a tuple $\pddlAction = \langle \mathit{name}(\pddlAction), \mathit{pre}(\pddlAction), \mathit{eff}(\pddlAction)\rangle$.
Here, $\mathit{name}(\pddlAction)$ is the name of action $\pddlAction$ and $\mathit{pre}(\pddlAction)$ is its precondition,  a first-order logical expression over \pddlPredicates\ which must be true for $\pddlAction$ to be executable. Effect $\mathit{eff}(\pddlAction)$ instead defines what
happens when $\pddlAction$ is executed. Action schemas can then be grounded in the same way as predicates.

In a problem specification $\pddlProblem = \langle \pddlObjects, \pddlInit, \pddlGoal \rangle$, the set \pddlObjects\ consists of objects $\pddlObject_k = \langle n, t\rangle$, where $n$ is the object name and $t \in \pddlTypes$
is its type. The initial state \pddlInit\ is a function that assigns a Boolean value to all propositions. Finally, $G$ defines which states are goal states via a logical expression over \pddlPredicates\ and \pddlObjects.

A plan $\pi$ is then a sequence of $m$ grounded actions $\pi = \langle a_1, \ldots, a_m \rangle$ such that these can be applied in sequence given their preconditions and that doing so transforms state $I$ into one satisfying $G$.

\subsection{Prompting Large Language Models}
LLMs are commonly controlled via descriptive inputs, so"-called \emph{prompts} \cite{gpt3}. We present the prompting techniques used in \name.
\emph{Few-shot prompting} entails adding examples of the desired behavior to the prompt and has been shown to allow LLMs to adapt to various tasks without fine-tuning \cite{gpt3}. \emph{Chain-of-thought reasoning} (CoT) encourages the LLM to reason and to analyze the task step by step, leading to improvements even without examples \cite{chainofthought, ZeroShotChainOfThought}.

\section{\name}

\name is to our knowledge the first complete assistive PDDL modeling system based on natural language, and by extension the first offline domain"-independent natural language planning tool. The only input \name needs is a task description in natural language which it then analyzes using an LLM, generating PDDL domain and problem files.
The generated PDDL files can then be given to a domain modeler, saving them time and effort. In some cases, \name even directly finds a first plan during its modeling process. This signals to the user that the generated domain model is ready to be passed to a planner that may find better, or even optimal, plans or that the domain can be used for solving other tasks.
Such a separation of concerns uses %
the LLM for its strengths---language comprehension and general world knowledge---while the planning task itself---which LLMs have been shown to struggle with---is left to the planner.
Using \name results in a significant reduction in workload
when applying classical planning to new domains, opening up the field to a wider audience.
Notably, \name does this without requiring any domain"-specific adaptations and is capable of working from shorter and less structured task descriptions than previous systems, see Section~\ref{sec:related}.

\begin{figure*}[ht]
    \centering
    \includegraphics[width=\textwidth, keepaspectratio]{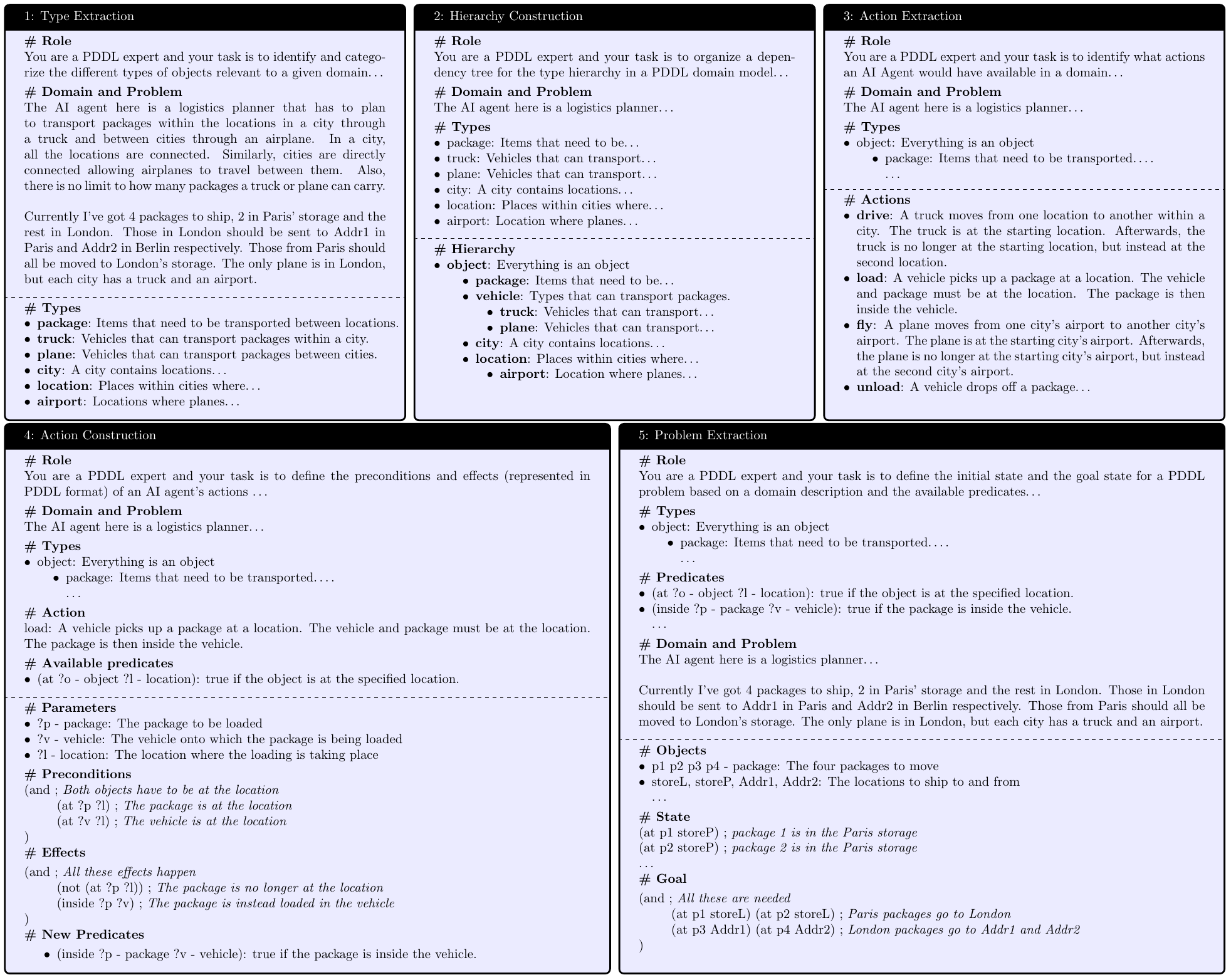}
    \caption{Example input-output pairs from the first five \name steps. For brevity, we omit portions of both.}%
    \label{fig:inputOutput}
\end{figure*}

The \name pipeline consists of six steps: Type Extraction, Hierarchy Construction, Action Extraction, Action Construction, Problem Extraction, and Planning.
We visualize the pipeline in Figure~\ref{fig:method} and show example input"-output pairs for each step are shown in Figure~\ref{fig:inputOutput}.
    For examples of the level of detail in the input, see Figure~\ref{fig:domains}.

We divided the \name pipeline into the six steps above to simplify the work for the LLM. Each step being limited in scope makes it easier for the LLM to perform it and allows for common failures to be easily identified and addressed through examples and questions within the feedback substeps.
The steps were ordered such that earlier results can inform later ones.

\subsection{Feedback and Validation}
All steps optionally use a feedback source---either an LLM or a human user.\footnote{We only evaluate \name using an LLM as feedback source.} If the feedback source finds issues with the initial attempt, the supplied feedback is used along with the initial output to obtain a revised version from the LLM. We only collect feedback once per step to avoid looping indefinitely and because a single feedback"-revision sequence is usually enough to correct most issues.

Steps 4--6 additionally use a suite of validation tools to ensure that the generated PDDL is valid. If validation fails, the error messages provided by these tools are used along with the latest version to regenerate a revised version. The validation tools used are VAL \cite{howey-long-icaps2003wscompetition}, Loki,\footnote{https://github.com/drexlerd/Loki} cpddl,\footnote{https://gitlab.com/danfis/cpddl} and a custom validator. The latter only detects errors in parentheses and keyword usage, but returns all flaws in a file at once and provides detailed suggestions for corrections. We chose a suite of validators as they give different feedback.

\subsection{Prompting}
All prompts in \name are three"-shot CoT prompts showcasing how to perform the task and the output format. For the main tasks, these prompts are illustrated in Figure~\ref{fig:inputOutput}. The revision substeps reuse this main prompt.

The feedback prompts additionally include a checklist of likely errors for the LLM to base its feedback on.
The first two exemplars showcase these common issues, while the third exemplar is entirely correct as to not overly bias the LLM towards always finding issues. This separation of generating and critiquing allows us to exemplify, and thereby correct, errors without ``confusing'' the main LLM.

All prompts were designed exclusively using non"-testing domains and with a different LLM (Llama-3.1 70B\footnote{https://ai.meta.com/blog/meta-llama-3-1}) to make the evaluation as unbiased as possible. As such, the prompts have not been adapted to the evaluation domains. Our full prompts, including the one for the baseline, are available online \cite{gestrin-et-al-zenodo2025} and the latest version is maintained at {\color{blue}\href{https://github.com/mrlab-ai/NL2Plan}{https://github.com/mrlab-ai/NL2Plan}}.

\subsection{\name Steps}
We now describe the six \name steps in more detail. %

\subsubsection{Step 1: Type Extraction}\hfill\\[0.2em]
In the first step, we use the LLM to extract the relevant types from the task description. The prompt does this by instructing the LLM to list each mentioned object and to identify its type. The resulting types are then collated and described, resulting in a list of types with descriptions.
The Logistics IPC domain, %
for example, requires packages, trucks, airplanes, locations, airports, and cities. The package type could be described as ``package: Items that need to be transported between locations''.

\subsubsection{Step 2: Hierarchy Construction}\hfill\\[0.2em]
The LLM is then used to organize the types into an inheritance tree by following a prompt that analyzes how the types are likely to be used in a PDDL model and then organizing them based on this usage. As an example, it could identify that both airplanes and trucks are used to transport packages and model them as subtypes of a \type{Vehicle} type. %

\subsubsection{Step 3: Action Extraction}\hfill\\[0.2em]
Next, we use the LLM to generate a list of actions based on the task description and the identified types. The prompt here encourages the LLM to generate short example action sequences that the generated model should support, such as ``Truck1 picks up package1 at the Paris storage, drives to Paris' airport, and unloads package1''. The LLM then identifies the actions used for these sequences, for example \action{load}, \action{drive}, and \action{unload}. Lastly, the found actions are expanded with natural language descriptions outlining their preconditions and effects. The \action{load} action for instance could be described as: ``load: A vehicle picks up a package at a location. Both the vehicle and package must be present at the location. Once loaded, the package is inside the vehicle.''

\begin{figure*}[!ht]
    \centering
    \includegraphics[width=\textwidth, keepaspectratio]{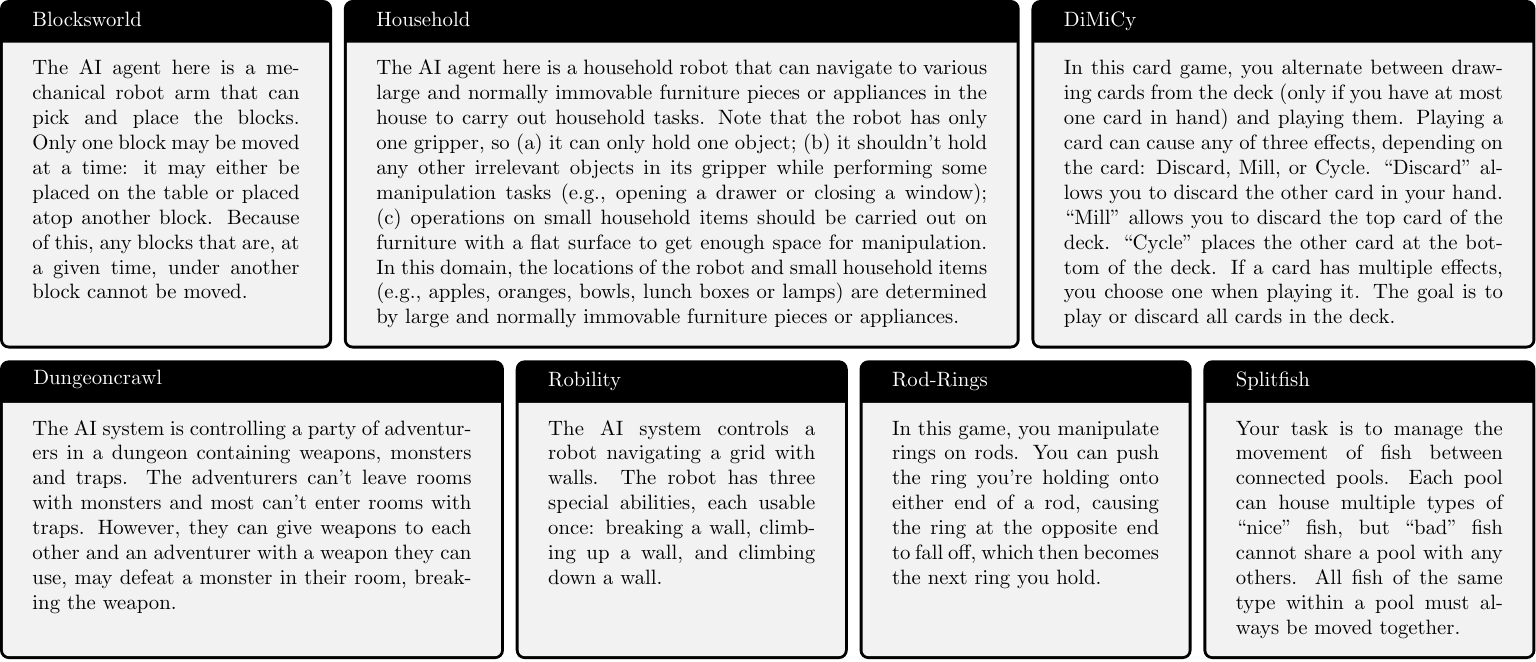} %
    \caption{Exact domain descriptions as given to \name and the baseline. For \blocksworld{} and \household{} the descriptions are shortened versions of those from \citet{llmConstructWorldModels}, the other domains are novel with descriptions written by us.
}
    \label{fig:domains}
\end{figure*}

\subsubsection{Step 4: Action Construction}\hfill\\[0.2em]
Building on the approach by \citet{llmConstructWorldModels}, we then formalize the PDDL domain one action at a time, simultaneously generating predicates. For differences to their work, see Section \ref{sec:differences}.
When constructing an action, the LLM is given the action name and description, as well as the types and predicates generated so far. The LLM then formulates the parameters, preconditions and effects of the action. It does so by first reasoning about these in detailed natural language and then writing the code. Each action is then individually validated by the validation suite and revised if the PDDL is incorrect. We allow for ADL syntax, which includes quantified and conditional effects, with negation, disjunction, and quantification in any conditions.

While generating an action, the LLM may define new predicates that are then added to the domain and passed on to future action synthetizations. These passed predicates are the only form of communication between the action"-generating LLM instances. Dynamically generating predicates this way means that the set of predicates does not need to be predefined by a user, which would require expertise and time, or an LLM, which is likely to make mistakes with such a complex job. Instead, the LLM can ``discover'' which predicates are needed as it models the domain. We also automatically generalize predicates, for example, changing \verb|(in ?p - package ?t - truck)| to \verb|(in ?p - package ?v - vehicle)| when the LLM later attempts to use this predicate for an airplane. This generalization replaces the type of the misused parameter with the lowest common ancestor of the current and attempted types.

Once all actions are defined, we have an entire PDDL domain file. The feedback source, either an LLM or a human, then checks this domain file for any issues, such as predicates that are used inconsistently or semantically invalid actions. If any issues are found, the feedback source provides suggestions on the action level, and the flawed actions are individually revised accordingly.

Finally, we clean up the domain file by removing any predicates and types  that are no longer used. This is done to ensure that the domain is as concise as possible, reducing the difficulty of Step~5.
At this point, we are done generating the PDDL domain file and turn to the PDDL problem file.

\subsubsection{Step 5: Problem Extraction}\hfill\\[0.2em]
To formalize the problem file, the LLM uses the type hierarchy from Step~2 and the predicates generated in Step~4 to describe the objects, their initial state, and the goal criteria. %

The full problem file is generated in one LLM call while interleaving natural language reasoning about the task with PDDL code. It is then validated section"-by"-section (first the objects, then the initial state, and finally the goal) by the validation suite. If errors are found, they are used to revise the problem file.

After the problem file has been generated in a syntactically correct manner, the feedback source checks the problem file for semantic issues. If any are found, the feedback source provides suggestions and the problem is revised accordingly.

\subsubsection{Step 6: Planning}\hfill\\[0.2em]
Lastly, the joint domain and problem files are checked by the validation suite.
Note, however, that initially errors can only occur
if either Step~4 or Step~5 ran out of attempts before correcting all errors.
Should the files be incorrect, the associated error messages are used by the LLM for revisions.
If they are correct, the files are passed to a classical planner and any plan found
is passed to a feedback substep which validates that said plan correctly aligns with the task description. If so, the plan is returned to the user. Otherwise, the feedback source identifies the flaws and suggests corrections which are used to revise the model.

However, should the syntactically correct model prove unsolvable, we use an algorithm to identify a set of changes to the initial state that would make the problem solvable. For this, we use a variant of the approach from \inlinecite{goebelbecker-et-al-icaps2010} that adds a set of overly expensive actions which can arbitrarily change the atoms, but only be executed before the first regular action of the plan.
This results in a, likely semantically wrong, potential fix for the task. To get better suggestions, we also use the LLM to identify how each predicate should be initialized and use this information to inform the algorithm. E.g., the \verb|(at ?v - vehicle ?l - location)| predicate should only be initialized once for each vehicle, but could be repeated for the location. The resulting changes are then provided as a suggestion to the LLM
which revises the domain and problem, while attempting to adhere to the original task description.

We use an algorithmic approach, instead of an LLM feedback source, since determining the cause of unsolvability is complex and often beyond the capabilities of an LLM.
The algorithm from \inlinecite{goebelbecker-et-al-icaps2010} was selected for its simplicity and compatibility with the full range of PDDL syntax we support.

After the LLM has revised the unsolvable model, this step is repeated until a plan is found or 10 revisions are performed, including syntax revisions. If no plan is found within 10 revisions, \name stops and returns the current model to the user. Regardless, the result is a PDDL domain, problem, and possibly a plan which are then returned to the user.

\newcommand{\baseshort}[0]{Base}
\newcommand{\oursshort}[0]{Ours}
\newcommand{\wins}[1]{\textbf{#1}}
\newcommand{\na}[0]{--}

\begin{table*}[!htb]

    \begin{center} %
        \def\nl{\\[-0.9em]} %
        \renewcommand{\arraystretch}{0.95}
        \setlength{\tabcolsep}{4pt}
        \begin{tabularx}{\textwidth}{@{}rRRRRRRRRRRRRRR@{}}
                   & \multicolumn{2}{c}{\blocksworld} & \multicolumn{2}{c}{\household} & \multicolumn{2}{c}{\dimicy} & \multicolumn{2}{c}{\dungeon} & \multicolumn{2}{c}{\robility} & \multicolumn{2}{c}{\rodrings} & \multicolumn{2}{c}{\splitfish} \\ \nl

        \cmidrule(l){2-3} \cmidrule(l){4-5} \cmidrule(l){6-7} \cmidrule(l){8-9} \cmidrule(l){10-11} \cmidrule(l){12-13} \cmidrule(l){14-15}

        Task    & \baseshort{} & \oursshort & \baseshort{} & \oursshort & \baseshort{} & \oursshort & \baseshort{} & \oursshort & \baseshort{} & \oursshort & \baseshort{} & \oursshort & \baseshort{} & \oursshort \\ \hline \nl
        1 & \wins{0} & 2    & 4 & \wins{0}   & \na{} & \na{}     & 1 & 1             & \na{} & \na{}    & \na{} & \na{} & 0 & 0             \cr \nl
        2 & \wins{0} & 1    & 1 & 1          & \na{} & \na{}     & \na{} & \wins{1}  & \na{} & \wins{1} & \na{} & \na{} & 4 & 4             \cr \nl
        3 & \wins{0} & 5    & \na{} & \na{}  & \na{} & \na{}     & \na{} & \wins{1}  & \na{} & \na{}    & \na{} & \na{} & \wins{3} & \na{}  \cr \nl
        4 & 1 & \wins{0}    & \na{} & \na{}  & \na{} & \na{}     & 3 & \wins{1}      & \na{} & \wins{3} & \na{} & \na{} & \na{} & \na{}     \cr \nl
        5 & \wins{3} & 4    & \na{} & \na{}  & \na{} & \na{}     & \na{} & \wins{1}  & \na{} & \na{}    & \na{} & \na{} & \na{} & \na{}     \cr \nl
        6 & \wins{0} & 2    & \na{} & \na{}  & \na{} & \na{}     & \na{} & \wins{3}  & \na{} & \na{}    & \na{} & \na{} & 3 & \wins{0}      \cr
        \bottomrule
        \end{tabularx}
        {
        \caption{The number of manual corrections needed for \baseline{} (\baseshort) and \name (\oursshort) to correctly model the specified tasks. We stop the evaluation if more than 5 changes would be required and report these cases as ``\na''. Lower numbers are better and tasks with higher IDs are generally more complex.}
        \label{tab:results}
        }
    \end{center}
\end{table*}

\section{Experiments}
We now describe our experiments for evaluating \name.

\subsection{Baseline}

\name is the first tool which can generate entire PDDL domains and problems from natural language descriptions.
\inlinecite{smirnov2024generatingconsistentpddldomains}, who developed their approach concurrently with \name, can work on the same input, but do not make their code available.
All other related tools require either further information or in-domain examples, see Section~\ref{sec:related}.
Since our focus is on methods for supporting domain modeling, approaches which ask LLMs to generate plans directly are not suitable baselines either.

For these reasons, we introduce a new baseline, called \baseline, which uses a zero"-shot prompted LLM to directly generate both the PDDL domain and problem file. The generated files are then validated by the same validation suite as \name and any errors are returned to the LLM for revisions. Once valid, the generated files are passed to a classical planner which solves the modeled task. This baseline retains the verifiability and
domain"-independent
properties of \name, while %
being simpler than the \name pipeline but stronger than directly asking an LLM---as a user would likely do.

\subsection{Domains and Problems}

To align with related work \egcite{RAPReasoningWithLanguageModelIsPlanning, kambhampatiPositionLLMsCant2024,zeroShotPlanners},
we evaluate \name on two commonly used domains, \blocksworld{} \cite{slaney-thiebaux-aij2001} and \household{} \cite{llmConstructWorldModels}. However, when researching with LLMs, there is the risk of data contamination. With well-established domains, there is a possibility that the LLM has trained on them, leading to potentially misleading evaluation results. To mitigate this issue, we created five completely new domains. These domains were selected for their concise descriptions, their feasibility for modeling in PDDL without reliance on numerics or other \name-unsupported features, and their ability to present non-trivial planning and modeling challenges. Together, they encompass a diverse range of problem types. Figure~\ref{fig:domains} shows the full-length descriptions of all seven domains.
For each domain, we generate 6 problems of varying difficulty: 2 ``short'' tasks (1 and 2) with optimal plans of 4--6 actions, 2 ``medium'' tasks (3 and 4) requiring 8--10 actions and 2 ``long'' (5 and 6) tasks requiring 12--14 actions. Plan length indirectly correlates with larger problems and more required actions, and are therefore usually more challenging to model.

\subsection{Experiment Setup}
For the classical planner, we use the first iteration of LAMA \cite{richter-westphal-jair2010}, implemented within Fast Downward \cite{helmert-jair2006}.
For the LLM, we use OpenAI's GPT models,
namely GPT"-4o"-2024"-08"-06, a commonly used model with a high performance vs.\ cost ratio.
While we use a temperature of 0 and specify a seed, GPT-4o is still not deterministic. As all experiments are evaluated manually, we prioritize trying the approaches on multiple tasks rather than running each task multiple times.
We only consider a single LLM, since our goal is not to compare LLMs, but to evaluate how a given LLM benefits from the \name{} pipeline when asked to model PDDL tasks. %

\subsection{Evaluation}
We manually evaluate the generated domains, problems, and plans for each method and task. We compare these to the natural language task descriptions---that is, the intent of the tasks---rather than directly to our own PDDL reference models. We choose this approach since there are many correct ways to model each task and therefore no single ground truth exists to compare to.
As we do not enforce a fixed set of predicates or actions, we cannot use any automatic evaluation as is done in some related work \egcite{stein2024autoplanbench, Oswald_Srinivas_Kokel_Lee_Katz_Sohrabi_2024}.
    Instead, we count the number of changes required to make the generated models semantically solve the task. For example: adding a predicate to a precondition, replacing a predicate with another in an effect, or removing a grounded predicate from the initial state.

\section{Results}
We now present the experiment results and their analysis. The detailed results are shown in Table~\ref{tab:results}. Summarizing these, while excluding Blocksworld due to likely data contamination, we see that \name consistently solves more tasks for the same maximum number of manual changes:
\begin{center}
\begin{tabular}{@{}rrr@{}}
\text{\# Changes} & \baseline{} \text{ solves} & \name{} \text{ solves} \\
\midrule
0 & 2.8\% & 8.3\% \\
$\le$1 & 8.3\% & 27.8\% \\
$\le$3 & 16.7\% & 33.3\% \\
$\le$5 & 22.2\% & 36.1\% \\
\bottomrule
\end{tabular}
\end{center}
Note that tasks requiring 0 changes would be solved perfectly by passing the task model to a planner, meaning that NL2Plan solves 260\% more unseen tasks than the baseline.

\subsection{Per-Domain Results}
We now analyze the performance of \baseline{} and \name per domain. For each domain we provide a summary of the results, highlighting the most important aspects.

\paragraph{\blocksworld}
    In \blocksworld, \baseline{} clearly shows that GPT-4o has prior exposure to the domain. It correctly identifies \blocksworld by name and follows standard predicate and action conventions with minimal naming variations. In contrast, the steps in \name appear to break up these conventions, usually (5/6 tasks) resulting in models that allow for multiple arms and tables with a joint \action{pick-up} action. However, this joint action would often (4/6 tasks) require a conditional adjustment to set the \predicate{clear} predicate correctly.%

\paragraph{\household}
In \household{}, both approaches have some difficulties with the intricate types. PDDL-0 never uses subtypes, though it uses the \keyword{either} keyword twice, which contributes to the model failures. Conversely, \name routinely uses subtypes and usually does so correctly. Each approach also has its own common mistakes: for \baseline{} it is being able to act when holding an object, while for \name it is not being able walk while carrying something.

\paragraph{\dimicy}
In \dimicy{}, neither method successfully models the domain. Both are unable to model deck order, at most using predicates for the top and bottom of the deck. Additionally, \baseline{} struggles with the action modeling, often only including a single \action{play} action that applies all effects.

\paragraph{\dungeon}
    Modeling \dungeon correctly requires that movement checks for traps at the destination, but for monsters at the origin. \baseline{} always mixes up the latter and \name does so for 4/6 tasks. \baseline{} sometimes (2/6 tasks) also fails to model traps, allowing any armed adventurer to bypass them. For connectivity, \name accurately represents bidirectionality, whereas \baseline{} does so inconsistently. \baseline{} omits the \action{pick-up} action in three needed cases, while \name only once excludes \action{give} when unnecessary. \name's only major mistake is allowing broken weapons to be used in one task, while \baseline{} has initialization errors in dungeon directionality and weapon placement.%
\paragraph{\robility}
For \robility{}, both approaches often put walls between two locations instead of at exactly once location. \baseline{} also commonly makes mistakes in connectivity: not including it at all, defining it in one direction, or missing connections. \name always models this correctly. %

\paragraph{\rodrings}
In \rodrings{},
neither approach correctly manages to model the order of the rings on the rods. Additionally, they usually model replacing an edge ring rather than pushing off the opposite one.
\paragraph{\splitfish}
\splitfish{} is the only domain after \blocksworld{} where \baseline{} performs better than \name. \name usually models each starting fish as an individual object rather than reusing the same object for multiple locations. \baseline{} also does this, but in fewer tasks. Likely, this is caused by \name's exemplars never reusing objects in this manner, and as such this a case where \name actually suffers from using three"-shot prompting. \baseline{} models the tasks as uni"-directional in most cases, however some tasks remain solvable even in this form.

\begin{table}[tb]
    \centering
        \renewcommand{\arraystretch}{0.95}
        \setlength{\tabcolsep}{4pt}
        \begin{tabular}{@{}lrrrrrr@{}}
                    &Step~1 &Step~2 &Step~3 &Step~4 &Step~5 &Step~6
            \\
            \midrule
            Good    & 16    & 3     & 7     & 15    & 11    & 17\\

            Mixed   & 0     & 1     & 0     & 3     & 0     & 3\\

            Bad     & 0     & 0     & 0     & 0     & 2     & 3\\
            \bottomrule
        \end{tabular}
        \caption{Number of tasks for which the \name LLM"-feedback gave advice, separated by step. The feedback is classified as ``Good'', when it suggests an improvement, ``Bad'', when it would make the result worse, and ``Mixed'', when it contains both good and bad elements.}
        \label{tab:feedback}
\end{table}
\subsection{Validation, Feedback and Suggestions}
    Validation is applied in both \name and \baseline{} to ensure the generated PDDL files are syntactically correct. Thanks to GPT-4o’s capabilities, few errors occur and they are often fixed after one validator message. \baseline{} retains syntax errors in 6 tasks, while \name only has 3 invalid tasks. The most common issues are mismatched types, especially in \household, and missing action parameters.

    The feedback substeps in \name, which aim to identify semantic errors using checklists, yield a high ratio of useful advice (69 good vs.\ 5 bad), as shown in Table~\ref{tab:feedback}. While the LLM"-based feedback sometimes misses some errors and although occasional misinterpretations occur---such as misidentifying a valid plan or misadvising on domain-specific actions---the feedback is clearly a net positive. For Step~6, however, the revision LLM sometimes struggles to correct identified plan-level errors, as troubleshooting PDDL can be complex.

    During Step~6, we also have the algorithmic suggestion substep to aid with unsolvability. As it only suggests initial state changes, this biases the revision LLM away from altering the more-often flawed domain. However, without any guidance the LLM would likely consider both PDDL files valid, and even with this bias the revision in several cases correctly links the suggested initial state changes to flawed actions. Regardless, a suggestion system also capable of suggesting changes to the domain file would likely improve the overall performance of \name.

\subsection{Quantifiers versus Predicates}
Across all domains, we find that both approaches perform better when using existential or universal quantifiers rather than predicates when possible. For instance, in \blocksworld, this means using \predicate{(not (exists ?b2) (on ?b2 ?b))} instead of \predicate{(clear ?b)}. While a human modeler would often use a single predicate, a quantifier does not require the LLM to ``remember'' to update the predicate later which seems to reduce the risk of errors.

\subsection{Token Usage and Cost}
The following table shows the average token usage and associated cost per task:

\begin{center}
    \setlength{\tabcolsep}{10pt}
    \begin{tabular}{@{}lrrr@{}}
        Method & Input & Output & Cost \\
        \midrule
        \baseline   & 4369 & 1660 & \$0.03\\
        \name     & 154609 & 17965 & \$0.57\\
        \bottomrule
    \end{tabular}
\end{center}
\noindent
The cost of \name is noticeably higher than that for \baseline{}, roughly by a factor of 19. The increased token usage is primarily due to the far longer prompts and %
the several LLM calls per task in \name. We did not optimize \name for token usage, and it is likely that many prompts could be shortened without losing quality. However, we consider the increased cost to not be prohibitive if it leads to a better model, thereby saving time for the user. %

\section{Related Work}
\label{sec:related}

Recent advances in LLMs have enabled planning methods that work directly on natural language problem descriptions \cite{planningAbilityLLM}. However, these often struggle with long-term action effects and physical constraints, making them less scalable for extended tasks \cite{planbench, stein2024autoplanbench, llmPlusP}. \emph{Online} LLM-based planners \egcite{ReAct,reflexion} aim to address this by interleaving action selection with environmental feedback. In contrast, \emph{offline} reasoners \egcite{chainofthought,treeOfThoughts,selfEvalGuidedBeamSearch,RAPReasoningWithLanguageModelIsPlanning} focus on high"-quality deduction with the planners being domain"-dependent \egcite{RAPReasoningWithLanguageModelIsPlanning,zeroShotPlanners}. Recently, large reasoning models (LRMs) have emerged \cite{OpenAI2024LearningReasonLLMs}, outperforming LLMs in planning tasks but remaining costly and opaque \cite{valmeekam2024llmscantplanlrms}. Despite their accessibility, LLM-based planners lack correctness guarantees without domain-specific adaptations \cite{kambhampatiPositionLLMsCant2024}.

Due to the complementary strengths of PDDL"- and LLM"-based methods, recent work mixes the two approaches in various ways. For example, LLMs have been used to generate parts of the PDDL descriptions \egcite{llmPlusP, dynamicPlanningLLM, llmConstructWorldModels, LLMsOutOfDistribution, lyu-etal-2023-faithful, Oswald_Srinivas_Kokel_Lee_Katz_Sohrabi_2024, huang2024planningdarkllmsymbolicplanning, mahdavi2024leveraging, zhu2024language}, to initialize classical planners \cite{planningAbilityLLM}, and to solve PDDL-specified problems \cite{pddlPlanningPretrainedLLMs,generalPlanPddl, plansformer, Oswald_Srinivas_Kokel_Lee_Katz_Sohrabi_2024, huang2024planningdarkllmsymbolicplanning}. Conversely, PDDL descriptions have been used to automatically validate LLM"-made plans \cite{planningAbilityLLM,kambhampatiPositionLLMsCant2024}. Each method focuses on generating parts of the PDDL, such as just the actions \cite{Oswald_Srinivas_Kokel_Lee_Katz_Sohrabi_2024, huang2024planningdarkllmsymbolicplanning}, the actions and predicates \cite{llmConstructWorldModels}, just the goal state \cite{lyu-etal-2023-faithful, xie2023translating}, or both the initial and goal state \cite{llmPlusP, dynamicPlanningLLM, LLMsOutOfDistribution}. These methods additionally require structured or domain-dependent inputs, such as other parts of the PDDL and examples from the current domain \egcite{llmPlusP, dynamicPlanningLLM, LLMsOutOfDistribution, lyu-etal-2023-faithful}, explicit and structured descriptions of all types and actions \cite{llmConstructWorldModels, Oswald_Srinivas_Kokel_Lee_Katz_Sohrabi_2024, huang2024planningdarkllmsymbolicplanning}, or access to the environment \cite{mahdavi2024leveraging, zhu2024language}.

We provide a brief overview of the above domain generation approaches which do not require environment interactions.
Like \name, \inlinecite{llmConstructWorldModels} generate actions sequentially with interleaved predicate creation, though without our feedback structures and repair steps.
\inlinecite{huang2024planningdarkllmsymbolicplanning} employ an LLM configured for variance to generate multiple candidate actions, filtering and recombining them based on similarity within an embedding space. While this method is computationally expensive, it allows users to select between possible recombinations and in their experiments outperforms the work by \inlinecite{llmConstructWorldModels}.
\inlinecite{Oswald_Srinivas_Kokel_Lee_Katz_Sohrabi_2024} instead focus on the evaluation: they use a 1-shot prompt to generate a drop-in replacement for a single action, with the exemplar being of another action from the same domain. Doing so, they can automatically evaluate the PDDL writing capabilities of LLMs.
The latter two require that the predicates are predefined and all three require that the actions are explicitly provided individually.

Concurrently to our work on \name, \inlinecite{smirnov2024generatingconsistentpddldomains} introduced another approach for generating PDDL domains and problems from natural language. However, they do not provide any source code, prompts, nor evaluation data. Furthermore, they do not release the domain or task descriptions and the only example given is ``make a pizza''. As such, it is impossible to judge their level of input detail---whether it is ``natural'' as in our work or heavily structured and explicit as in others \egcite{stein2024autoplanbench}. Moreover, their evaluation is limited to verifying whether the generated PDDL can be parsed and solved, which, as we demonstrate, does not necessarily imply that the natural language task is solved correctly. Although the authors acknowledge this limitation, they do not conduct any semantic-level analysis and only consider one novel domain. Last, it is also unclear whether their prompts were developed on the test domains. In contrast, we fully open-source our work, conduct evaluations across a diverse set of unseen domains, perform manual semantic analysis of the generated PDDL, and use a pipeline developed only on non"-testing domains. In general, our focus is on generating both syntactically and semantically correct models, while \citeauthor{smirnov2024generatingconsistentpddldomains} focus exclusively on the former.

There are also non"-LLM"-based methods to generate PDDL autonomously, but these generally require examples of plans, so"-called traces, or environment interactions, making them more difficult to apply \cite{arora-et-al-ker2018}.

\section{Differences to Prior and Concurrent Work}
\label{sec:differences}
While the Action Construction step of \name is based on work by \citet{llmConstructWorldModels}, we allow for non-STRIPS actions, change the prompt methodology and add automatic predicate generalization, which can result in clearer and more concise domains. Additionally, Steps 1--3 of \name relieve the user from having to explicitly define the types and the actions, increasing ease of use and applicability. Lastly, the new feedback substep both reduces costs, since only flawed actions are revised, and allows for complex, semantic issues to be corrected.
    \citet{llmConstructWorldModels} also did not attempt to create problem files.

The Problem Extraction step is inspired by \citet{llmPlusP}, \citet{LLMsOutOfDistribution} and other similar approaches. In addition to not requiring handcrafted PDDL domains due to Steps 1--4, \name is the first system that operates on ambiguous, free"-form text as input. In contrast, previous approaches were only used for structured, automatically generated descriptions. The addition of a feedback substep is also novel for this part of the pipeline. Lastly, \name offers the first robust domain"-agnostic variant, with most previous work including in"-domain examples of how to model problems or domain"-specific adaptations. While \citet{llmPlusP} also introduced a zero-shot version of LLM+P, requiring only a handcrafted domain file, it failed to solve any task.

Comparing our pipeline to \inlinecite{smirnov2024generatingconsistentpddldomains}, we observe both similarities and key differences. Their Step~1, akin to our Step~3, selects actions using LLM"-generated plans; however, they ask the LLM to solve the full original problem rather than simplified hypothetical ones, as we do. We hypothesize that our variant generates more diverse action sets, which may lead to considering actions that are unneeded for the task at hand, but required for other tasks from the same domain. In their Step~2, \citeauthor{smirnov2024generatingconsistentpddldomains} generate the domain and problem using a JSON structure, aiming to reduce syntax errors. However, their reported error rate is significantly higher than ours, though we note that they use GPT-4 and we use GPT-4o. Our separation across multiple more fine"-grained steps also offers the possibility to better support the system and for validation at each step if desired. Finally, they perform an algorithmic reachability analysis to identify actions that can never be executed. While this could be incorporated into \name, unreachable actions are not necessarily problematic, since not all actions may be required for every instance. Moreover, their algorithm, as most domain repair approaches, is constrained to ``lower-level'' PDDL (STRIPS with negation).

\section{Conclusions and Future Work}
\name is able to generate PDDL files for a variety of domains and problems, requiring zero or one predicate changes to correctly model $27.8\%$ of tasks, outperforming a direct approach only combining an LLM with a validation suite. Additionally, \name fully solved 260\% more tasks than said baseline. The separation of the modeling process clearly helps \name correctly model tasks, making it rarely forget to include necessary actions while also modeling more general cases. The feedback substeps similarly improve the quality of the generated models, identifying several semantic errors that would otherwise have been included while rarely giving bad advice. As such, \name is a useful assistive tool for generating PDDL files from natural language descriptions, while also being a step towards a fully automated natural language planner which provides guarantees and interpretability.

\bibliography{abbrv-short,literatur,extra,crossref-short}

\end{document}